\documentclass[10pt]{article}
\IfFileExists{tmlr.sty}{\usepackage[preprint]{tmlr}}{\usepackage[margin=1in]{geometry}\usepackage[round]{natbib}}
\usepackage[T1]{fontenc}
\usepackage[utf8]{inputenc}
\usepackage{graphicx,booktabs,amsmath,hyperref,url}
\hypersetup{hidelinks}

\newcommand{\NmBrCbefore}{0.0019}
\newcommand{\NmBrCafter}{0.011}
\newcommand{\NmBrCfbefore}{7}
\newcommand{\NmBrCfafter}{12}
\newcommand{\NmBrCnbefore}{3{,}713}
\newcommand{\NmBrCnafter}{1{,}063}

\newcommand{\NmBrCp}{$1.5\times10^{-4}$}

\newcommand{\NmBrBp}{0.56}

\newcommand{\NmBctrsteps}{8{,}000}
\newcommand{\NmBrBslicemin}{0.0037}
\newcommand{\NmBrBslicemax}{0.0049}

\newcommand{\NmCrCbefore}{0.0016}
\newcommand{\NmCrCafter}{0.011}

\newcommand{\NmCrCp}{$3.5\times10^{-5}$}

\newcommand{\NbErCbefore}{0.0082}
\newcommand{\NbErCafter}{0.018}
\newcommand{\NbErCp}{0.38}
\newcommand{\NbErCnafter}{56}
\newcommand{\NbEctrfaults}{17}
\newcommand{\NbEctrsteps}{4{,}408}
\newcommand{\Nfloorunloadseeds}{5}
\newcommand{\Nfloorunloadpasses}{3}
\newcommand{\Nfloorunloadsteps}{6{,}000}
\newcommand{\Nfloorunloadrangefaults}{0}

\newcommand{\Nfloorhotseeds}{3}
\newcommand{\Nfloorhotsteps}{4{,}800}
\newcommand{\Nfloorhotrangefaults}{1}

\newcommand{\Nfloorhothazard}{0.00021}

\newcommand{\NmBarmaseeds}{4}

\newcommand{\NmBarmdseeds}{20}

\newcommand{\NmBarmdctrseedshit}{20}

\newcommand{\NmBbdctrratio}{2.3$\times$}
\newcommand{\NmBbdctrp}{$7.6\times10^{-3}$}
\newcommand{\NmBbdrCp}{$8.9\times10^{-4}$}
\newcommand{\NmBbdrBp}{0.41}

\newcommand{\NmBbArCp}{$6.9\times10^{-3}$}

\newcommand{\NmBarmbrCrisep}{$2.9\times10^{-3}$}

\newcommand{\NmCarmAseeds}{20}
\newcommand{\NmCarmActr}{0}

\newcommand{\NmCarmActrseedshit}{0}
\newcommand{\NmCarmArC}{0.0039}
\newcommand{\NmCarmArB}{0.0045}
\newcommand{\NmCarmaseeds}{8}
\newcommand{\NmCarmactr}{0.0028}

\newcommand{\NmCarmactrseedshit}{6}
\newcommand{\NmCarmarC}{0.14}

\newcommand{\NmCarmbseeds}{20}
\newcommand{\NmCarmbctr}{0.019}

\newcommand{\NmCarmbctrseedshit}{20}
\newcommand{\NmCarmbrC}{0.0019}
\newcommand{\NmCarmbrB}{0.0052}
\newcommand{\NmCarmdseeds}{20}
\newcommand{\NmCarmdctr}{0.0071}

\newcommand{\NmCarmdctrseedshit}{20}
\newcommand{\NmCarmdrC}{0.0048}
\newcommand{\NmCarmdrB}{0.0055}
\newcommand{\NmCbdctrratio}{2.6$\times$}
\newcommand{\NmCbdctrp}{$2.1\times10^{-3}$}
\newcommand{\NmCbdrCp}{$6.0\times10^{-3}$}
\newcommand{\NmCbdrBp}{0.49}

\newcommand{\NmCbArCp}{0.031}
\newcommand{\NmCbArBp}{0.71}
\newcommand{\NmCdArCpworse}{0.31}
\newcommand{\NmCarmbrCrisep}{$3.0\times10^{-3}$}
\newcommand{\NmCarmarCfirstf}{8}
\newcommand{\NmCarmarCfirstn}{56}
\newcommand{\NmCarmarCfirst}{0.14}
\newcommand{\NmCarmArCfirstf}{0}

\newcommand{\NmCarmbctrfirstmedian}{18}
\newcommand{\NmCarmbctrfirstminusB}{14}
\newcommand{\NmCarmdctrfirstmedian}{114}

\newcommand{\NmCarmAtokenssBAA}{5{,}860.5}
\newcommand{\NmCarmbtokenssBAA}{6{,}250}
\newcommand{\NmCarmdtokenssBAA}{6{,}668}
\newcommand{\NcutEb}{31}

\newcommand{\NcutIb}{147}

\newcommand{\NcutBEb}{173}

\newcommand{\NcutDCb}{100}

\newcommand{\NexplmedianEb}{31.5}
\newcommand{\NexplmedianIb}{147.5}
\newcommand{\NexplmedianBEb}{173}
\newcommand{\NexplmedianDCb}{100.5}
\newcommand{\Ncutoffroundingchanges}{0}

\newcommand{\NsealedIbrCbefore}{0.00070}
\newcommand{\NsealedIbrCafter}{0.0096}

\newcommand{\NsealedIbrCp}{$3.8\times10^{-6}$}

\newcommand{\NsealedBEbrCexpectedhA}{2.8}

\newcommand{\NsealedDCbrCexpectedhA}{0}

\newcommand{\NcutBHb}{0}

\newcommand{\NaltEbinvalid}{20}
\newcommand{\NaltEbtotal}{20}
\newcommand{\Ncutcommon}{173}
\newcommand{\NffownEbh}{0.012}

\newcommand{\NffownIbh}{0.0096}

\newcommand{\NffownBEbh}{0.0016}

\newcommand{\NffownDCbh}{0.00034}

\newcommand{\NffownIbBEbratio}{6.2$\times$}

\newcommand{\NffownIbDCbratio}{28$\times$}

\newcommand{\NffcomEbh}{0.015}
\newcommand{\NffcomEbbound}{0.072}
\newcommand{\NffcomEbarisque}{1}
\newcommand{\NffcomEbcensurees}{0}
\newcommand{\NffcomEbfautes}{1}
\newcommand{\NffcomEbpas}{66}
\newcommand{\NffcomIbh}{0.013}
\newcommand{\NffcomIbbound}{0.019}
\newcommand{\NffcomIbarisque}{18}
\newcommand{\NffcomIbcensurees}{0}
\newcommand{\NffcomIbfautes}{18}
\newcommand{\NffcomIbpas}{1{,}402}
\newcommand{\NffcomBEbh}{0.0016}
\newcommand{\NffcomBEbbound}{0.0033}
\newcommand{\NffcomBEbarisque}{17}
\newcommand{\NffcomBEbcensurees}{12}
\newcommand{\NffcomBEbfautes}{5}
\newcommand{\NffcomBEbpas}{3{,}223}
\newcommand{\NffcomDCbh}{0.00046}
\newcommand{\NffcomDCbbound}{0.0014}
\newcommand{\NffcomDCbarisque}{20}
\newcommand{\NffcomDCbcensurees}{18}
\newcommand{\NffcomDCbfautes}{2}
\newcommand{\NffcomDCbpas}{4{,}344}
\newcommand{\NffcomIbBEbratio}{8.3$\times$}
\newcommand{\NffcomIbBEbp}{$2.7\times10^{-6}$}
\newcommand{\NffcomIbDCbratio}{28$\times$}
\newcommand{\NffcomIbDCbp}{$9.8\times10^{-10}$}

\newcommand{\NafcomEbh}{0.050}

\newcommand{\NafcomIbh}{0.018}

\newcommand{\NafcomBEbh}{0.0015}

\newcommand{\NafcomDCbh}{0.00044}

\newcommand{\NseedIbBEbp}{$2.2\times10^{-7}$}
\newcommand{\NseedIbDCbp}{$1.5\times10^{-8}$}
\newcommand{\NseedBEbDCbp}{0.057}

\newcommand{\NseednofaultIb}{0}
\newcommand{\NseednofaultBEb}{14}
\newcommand{\NseednofaultDCb}{18}

\newcommand{\NextIb}{2}

\newcommand{\NextBEb}{5}

\newcommand{\NextDCb}{11}
\newcommand{\NearlyDCb}{5}
\newcommand{\NfullDCb}{6}
\newcommand{\NextIbBEbp}{0.20}
\newcommand{\NextIbDCbp}{$2.9\times10^{-3}$}
\newcommand{\NcleanmBfbefore}{3}
\newcommand{\NcleanmBnbefore}{2{,}682}
\newcommand{\NcleanmBfafter}{5}
\newcommand{\NcleanmBnafter}{385}

\newcommand{\NcleanmCfbefore}{2}
\newcommand{\NcleanmCnbefore}{2{,}612}
\newcommand{\NcleanmCfafter}{5}
\newcommand{\NcleanmCnafter}{414}

\newcommand{\NrenmBcBp}{$1.5\times10^{-4}$}

\newcommand{\NrenmBarmArBf}{53}
\newcommand{\NrenmBarmArBn}{7{,}560}
\newcommand{\NrenmBarmActrf}{0}
\newcommand{\NrenmBarmActrn}{8{,}000}
\newcommand{\NrenmBarmAhidden}{0}

\newcommand{\NrenmBarmarBf}{11}
\newcommand{\NrenmBarmarBn}{1{,}520}
\newcommand{\NrenmBarmactrf}{673}
\newcommand{\NrenmBarmactrn}{1{,}600}
\newcommand{\NrenmBarmahidden}{8}
\newcommand{\NrenmBarmarBratio}{1.2}
\newcommand{\NrenmBarmarBlo}{1.0}
\newcommand{\NrenmBarmarBhi}{2.5}

\newcommand{\NrenmBarmarBstate}{not conclusive}

\newcommand{\NrenmBarmbrBf}{45}
\newcommand{\NrenmBarmbrBn}{7{,}597}
\newcommand{\NrenmBarmbctrf}{6{,}682}
\newcommand{\NrenmBarmbctrn}{8{,}000}
\newcommand{\NrenmBarmbhidden}{37}
\newcommand{\NrenmBarmbrBratio}{0.84}
\newcommand{\NrenmBarmbrBlo}{0.52}
\newcommand{\NrenmBarmbrBhi}{1.4}

\newcommand{\NrenmBarmbrBstate}{upper bound < 2}

\newcommand{\NrenmBarmdrBf}{51}
\newcommand{\NrenmBarmdrBn}{7{,}591}
\newcommand{\NrenmBarmdctrf}{5{,}479}
\newcommand{\NrenmBarmdctrn}{8{,}000}
\newcommand{\NrenmBarmdhidden}{29}
\newcommand{\NrenmBarmdrBratio}{0.96}
\newcommand{\NrenmBarmdrBlo}{0.59}
\newcommand{\NrenmBarmdrBhi}{1.6}
\newcommand{\NrenmBarmdrBhipct}{62}
\newcommand{\NrenmBarmdrBstate}{upper bound < 2}
\newcommand{\NrenmCcBp}{$3.5\times10^{-5}$}
\newcommand{\NrenmCarmArB}{0.0077}
\newcommand{\NrenmCarmArBf}{58}
\newcommand{\NrenmCarmArBn}{7{,}543}
\newcommand{\NrenmCarmActrf}{0}
\newcommand{\NrenmCarmActrn}{8{,}000}
\newcommand{\NrenmCarmAhidden}{0}
\newcommand{\NrenmCarmarB}{0.0069}
\newcommand{\NrenmCarmarBf}{21}
\newcommand{\NrenmCarmarBn}{3{,}040}
\newcommand{\NrenmCarmactrf}{1{,}045}
\newcommand{\NrenmCarmactrn}{3{,}200}
\newcommand{\NrenmCarmahidden}{9}
\newcommand{\NrenmCarmarBratio}{1.9}
\newcommand{\NrenmCarmarBlo}{0.93}
\newcommand{\NrenmCarmarBhi}{7.5}

\newcommand{\NrenmCarmarBstate}{not conclusive}
\newcommand{\NrenmCarmbrB}{0.0057}
\newcommand{\NrenmCarmbrBf}{43}
\newcommand{\NrenmCarmbrBn}{7{,}593}
\newcommand{\NrenmCarmbctrf}{6{,}732}
\newcommand{\NrenmCarmbctrn}{8{,}000}
\newcommand{\NrenmCarmbhidden}{38}
\newcommand{\NrenmCarmbrBratio}{0.74}
\newcommand{\NrenmCarmbrBlo}{0.45}
\newcommand{\NrenmCarmbrBhi}{1.2}

\newcommand{\NrenmCarmbrBstate}{upper bound < 2}
\newcommand{\NrenmCarmdrB}{0.0058}
\newcommand{\NrenmCarmdrBf}{44}
\newcommand{\NrenmCarmdrBn}{7{,}584}
\newcommand{\NrenmCarmdctrf}{5{,}072}
\newcommand{\NrenmCarmdctrn}{8{,}000}
\newcommand{\NrenmCarmdhidden}{22}
\newcommand{\NrenmCarmdrBratio}{0.75}
\newcommand{\NrenmCarmdrBlo}{0.49}
\newcommand{\NrenmCarmdrBhi}{1.2}

\newcommand{\NrenmCarmdrBstate}{upper bound < 2}
\newcommand{\NrencCverdict}{holds}

\newcommand{\NrenmBarmafirstctrchanged}{1}

\newcommand{\NrenmBarmdfirstctrchanged}{1}

\newcommand{\NrenmCarmafirstctrchanged}{2}

\newcommand{\NrenmCarmdfirstctrchanged}{0}
\newcommand{\NrensizectrseedsEb}{14}

\newcommand{\NrensizectrseedsDCb}{2}
\newcommand{\NrensizectrerrDCb}{2}
\newcommand{\NshortmBfbefore}{7}
\newcommand{\NshortmBnbefore}{120}
\newcommand{\NshortmBfafter}{4}
\newcommand{\NshortmBnafter}{100}
\newcommand{\NfamBfbefore}{1}
\newcommand{\NfamBnbefore}{3{,}860}
\newcommand{\NfamBfafter}{39}
\newcommand{\NfamBnafter}{3{,}740}
\newcommand{\NdBFAmBfafter}{20}
\newcommand{\NdBFAmBnafter}{80}

\newcommand{\NdDAAmBfafter}{8}
\newcommand{\NdDAAmBnafter}{80}

\newcommand{\NrCtimemB}{8.9$\times$}
\newcommand{\NrCtimemBlo}{5.2}
\newcommand{\NrCtimemBhi}{22}
\newcommand{\NrBtimemB}{1.3$\times$}
\newcommand{\NrBtimemBlo}{0.82}
\newcommand{\NrBtimemBhi}{2.3}

\newcommand{\NinterdmB}{0.69}
\newcommand{\NinterdmBlo}{0.60}
\newcommand{\NinterdmBhi}{0.77}

\newcommand{\NshortmCfbefore}{6}
\newcommand{\NshortmCnbefore}{120}
\newcommand{\NshortmCfafter}{4}
\newcommand{\NshortmCnafter}{100}
\newcommand{\NfamCfbefore}{1}

\newcommand{\NfamCfafter}{56}

\newcommand{\NrCtimemC}{12$\times$}
\newcommand{\NrCtimemClo}{6.3}
\newcommand{\NrCtimemChi}{35}
\newcommand{\NrBtimemC}{1.5$\times$}
\newcommand{\NrBtimemClo}{1.0}
\newcommand{\NrBtimemChi}{2.6}
\newcommand{\NdArCmC}{1.2$\times$}
\newcommand{\NdArCmClo}{0.71}
\newcommand{\NdArCmChi}{2.1}
\newcommand{\NbArCmC}{0.23$\times$}
\newcommand{\NbArCmClo}{0.14}
\newcommand{\NbArCmChi}{0.36}
\newcommand{\NbdrCmC}{0.19$\times$}
\newcommand{\NbdrCmClo}{0.10}
\newcommand{\NbdrCmChi}{0.32}
\newcommand{\NinterdmC}{0.64}
\newcommand{\NinterdmClo}{0.53}
\newcommand{\NinterdmChi}{0.73}
\newcommand{\NinterdmCpr}{$9.5\times10^{-7}$}
\newcommand{\NseedscidmClo}{0.83}
\newcommand{\NseedsciAmChid}{0.17}
\newcommand{\NmcnemardmCpr}{$9.5\times10^{-7}$}

\newcommand{\Nvalmaxtokens}{24{,}401}
\newcommand{\Nvalnumctx}{40{,}960}
\newcommand{\Nvalepisodes}{212}

\newcommand{\NdatephaseAhazardcriterion}{2026-08-12 17:40}
\newcommand{\NdatephaseAprereg}{2026-08-12 16:21}
\newcommand{\NdatephaseAhazardseal}{2026-08-12 21:47}

\newcommand{\NdatephaseBprereg}{2026-08-12 22:41}
\newcommand{\NdatephaseBdata}{2026-08-12 23:58}
\newcommand{\Ndatearmaprereg}{2026-08-13 15:49}
\newcommand{\Ndatearmadata}{2026-08-14 16:12}
\newcommand{\Ndatesizemapprereg}{2026-08-15 17:58}
\newcommand{\Ndatesizemapdata}{2026-08-16 16:24}

\newcommand{\NpAhfbefore}{7}
\newcommand{\NpAhnbefore}{3{,}113}
\newcommand{\NpAhfafter}{9}
\newcommand{\NpAhnafter}{757}
\newcommand{\NpAhbefore}{0.0022}
\newcommand{\NpAhafter}{0.012}
\newcommand{\NpAhpr}{$1.2\times10^{-3}$}

\newcommand{\NcmpmBbratio}{0.85}
\newcommand{\NcmpmBblo}{0.52}
\newcommand{\NcmpmBbhi}{1.4}

\newcommand{\NcmpmBdratio}{0.96}
\newcommand{\NcmpmBdlo}{0.60}
\newcommand{\NcmpmBdhi}{1.6}

\newcommand{\NunobsmCarmApct}{0.75}

\newcommand{\NunobsmCarmbpct}{0.092}

\newcommand{\NunobsmCarmdpct}{0.21}

\newcommand{\NcmpmCsteps}{7{,}531}
\newcommand{\NcmpmCbratio}{0.74}
\newcommand{\NcmpmCblo}{0.45}
\newcommand{\NcmpmCbhi}{1.3}

\newcommand{\NcmpmCdratio}{0.76}
\newcommand{\NcmpmCdlo}{0.49}
\newcommand{\NcmpmCdhi}{1.2}

\newcommand{\NauditmBbArBpr}{$1.5\times10^{-3}$}
\newcommand{\NauditmCarmArB}{0.0043}
\newcommand{\NauditmCarmbrB}{0.00076}
\newcommand{\NauditmCarmdrB}{0.0022}
\newcommand{\NauditmCbArBpr}{$2.0\times10^{-4}$}
\newcommand{\NpIpontmaxecart}{0}
\newcommand{\NpIrAseeds}{20}

\newcommand{\NpIrApropage}{20}

\newcommand{\NpIrAicprolo}{0.83}
\newcommand{\NpIrAicprohi}{1.0}
\newcommand{\NpIrrseeds}{20}
\newcommand{\NpIrrrecale}{20}

\newcommand{\NpIrrdelaimedian}{1}
\newcommand{\NpIrrdelaimax}{2}
\newcommand{\NpIrAverdict}{pattern continuation supported}
\newcommand{\NpItestBpr}{$7.3\times10^{-12}$}
\newcommand{\NpIpairsdisc}{20}
\newcommand{\NpIpairsrev}{0}
\newcommand{\NpItestBpairedpr}{$9.5\times10^{-7}$}
\newcommand{\NpIlseeds}{20}

\newcommand{\NpIlratio}{0.91}
\newcommand{\NpIllo}{0.34}
\newcommand{\NpIlhi}{2.5}
\newcommand{\NpIlverdict}{not conclusive}
\newcommand{\NpItestCpr}{0.94}

\newcommand{\NpIholmtestC}{does not survive}
\newcommand{\NpIlhitb}{19}
\newcommand{\NpIlhitl}{19}

\newcommand{\NpIlfirstb}{18}
\newcommand{\NpIlfirstl}{15.5}
\newcommand{\NpIlrBratio}{1.0}
\newcommand{\NpIlrBlo}{0.59}
\newcommand{\NpIlrBhi}{2.0}
\newcommand{\NpIlfab}{1}
\newcommand{\NpIlfal}{7}
\newcommand{\NpIlfan}{3{,}700}
\newcommand{\NpIltokdiffpct}{0.14}
\newcommand{\NpBAseeds}{20}
\newcommand{\NpBAhitbare}{0}

\newcommand{\NpBArCbare}{0.20}

\newcommand{\NpBAhitd}{15}
\newcommand{\NpBApropd}{37}
\newcommand{\NpBArBd}{0.73}
\newcommand{\NpBArCd}{1.2}
\newcommand{\NpBAtokd}{6{,}655.5}
\newcommand{\NpBAlatd}{13}
\newcommand{\NpBAhitds}{0}
\newcommand{\NpBApropds}{0}
\newcommand{\NpBArBds}{0.73}
\newcommand{\NpBArCds}{0.25}
\newcommand{\NpBAtokds}{6{,}672}
\newcommand{\NpBAlatds}{14}
\newcommand{\NpBAhitdone}{15}
\newcommand{\NpBApropdone}{21}

\newcommand{\NpBAhitdn}{19}
\newcommand{\NpBApropdn}{43}

\newcommand{\NpBArCdn}{1.8}
\newcommand{\NpBAtokdn}{6{,}670.5}

\newcommand{\NpBArepratio}{0}
\newcommand{\NpBAreplo}{0}
\newcommand{\NpBArephi}{0}
\newcommand{\NpBArepverdict}{repairs}
\newcommand{\NpBAdmgrBratio}{1.0}
\newcommand{\NpBAdmgrBlo}{0.67}
\newcommand{\NpBAdmgrBhi}{1.6}
\newcommand{\NpBAdmgrCratio}{0.20}
\newcommand{\NpBAdmgrClo}{0.079}
\newcommand{\NpBAdmgrChi}{0.67}
\newcommand{\NpBAdmgverdict}{not conclusive}
\newcommand{\NpBAdoneana}{15}
\newcommand{\NpBAdonenonana}{5}
\newcommand{\NpBAdonerecale}{8}
\newcommand{\NpBAdonerechute}{4}
\newcommand{\NpBAdonelo}{0.078}
\newcommand{\NpBAdonehi}{0.55}
\newcommand{\NpBAdoneverdict}{not conclusive}

\newcommand{\NpBAdonedelaimax}{1}

\newcommand{\NpBAcatnonen}{5}
\newcommand{\NpBAcatnoneseeds}{1 (35), 6 (69), 14 (49), 16 (69), 18 (160)}
\newcommand{\NpBAcatdurablen}{4}
\newcommand{\NpBAcatdurableseeds}{7 (39), 8 (109), 9 (59), 17 (179)}
\newcommand{\NpBAcatrelapsen}{4}
\newcommand{\NpBAcatrelapseseeds}{3 (159), 4 (139), 10 (160), 20 (39)}
\newcommand{\NpBAcatbriefn}{1}
\newcommand{\NpBAcatbriefseeds}{15 (17)}
\newcommand{\NpBAcatlaten}{1}
\newcommand{\NpBAcatlateseeds}{5 (198)}
\newcommand{\NpBArelreclo}{0.16}
\newcommand{\NpBArelrechi}{0.84}
\newcommand{\NpBAdscplo}{0}
\newcommand{\NpBAdscphi}{0.17}
\newcommand{\NpBApairdisc}{15}
\newcommand{\NpBApairrev}{0}
\newcommand{\NpBApairpr}{$3.1\times10^{-5}$}

\newcommand{\NpBAddBerrmatch}{15}
\newcommand{\NpBAddBnone}{5}
\newcommand{\NpBAbatteryeps}{0}
\newcommand{\NpBAtBpr}{$6.5\times10^{-5}$}
\newcommand{\NpBAtCpr}{0.42}
\newcommand{\NpBAholmtB}{survives}
\newcommand{\NpBAholmtC}{does not survive}
\newcommand{\NpBAinvalidsleep}{2}
\newcommand{\NpIvalepisodes}{80}
\newcommand{\NpIvalexcluded}{0}
\newcommand{\NpBAvalepisodes}{80}
\newcommand{\NpBAvalexcluded}{0}
\newcommand{\NholmgbErCp}{no}
\newcommand{\NholmfbErCp}{no}
\newcommand{\NholmgextIbBEbp}{no}
\newcommand{\NholmfextIbBEbp}{no}
\newcommand{\NholmgextIbDCbp}{\textbf{no}}
\newcommand{\NholmfextIbDCbp}{yes}
\newcommand{\NholmgffcomIbBEbp}{yes}
\newcommand{\NholmfffcomIbBEbp}{yes}
\newcommand{\NholmgffcomIbDCbp}{yes}
\newcommand{\NholmfffcomIbDCbp}{yes}
\newcommand{\NholmgmBarmbrCrisep}{\textbf{no}}
\newcommand{\NholmfmBarmbrCrisep}{\textbf{no}}
\newcommand{\NholmgmBbArCp}{\textbf{no}}
\newcommand{\NholmfmBbArCp}{\textbf{no}}
\newcommand{\NholmgmBbdctrp}{\textbf{no}}
\newcommand{\NholmfmBbdctrp}{yes}
\newcommand{\NholmgmBbdrBp}{no}
\newcommand{\NholmfmBbdrBp}{no}
\newcommand{\NholmgmBbdrCp}{\textbf{no}}
\newcommand{\NholmfmBbdrCp}{yes}
\newcommand{\NholmgmBrBp}{no}
\newcommand{\NholmfmBrBp}{no}
\newcommand{\NholmgmBrCp}{yes}
\newcommand{\NholmfmBrCp}{yes}
\newcommand{\NholmgmCarmbrCrisep}{\textbf{no}}
\newcommand{\NholmfmCarmbrCrisep}{\textbf{no}}
\newcommand{\NholmgmCbArBp}{no}
\newcommand{\NholmfmCbArBp}{no}
\newcommand{\NholmgmCbArCp}{no}
\newcommand{\NholmfmCbArCp}{no}
\newcommand{\NholmgmCbdctrp}{\textbf{no}}
\newcommand{\NholmfmCbdctrp}{yes}
\newcommand{\NholmgmCbdrBp}{no}
\newcommand{\NholmfmCbdrBp}{no}
\newcommand{\NholmgmCbdrCp}{\textbf{no}}
\newcommand{\NholmfmCbdrCp}{\textbf{no}}
\newcommand{\NholmgmCdArCpworse}{no}
\newcommand{\NholmfmCdArCpworse}{no}
\newcommand{\NholmgmCrCp}{yes}
\newcommand{\NholmfmCrCp}{yes}
\newcommand{\NholmgsealedIbrCp}{yes}
\newcommand{\NholmfsealedIbrCp}{yes}
\newcommand{\NholmgseedBEbDCbp}{no}
\newcommand{\NholmfseedBEbDCbp}{no}
\newcommand{\NholmgseedIbBEbp}{yes}
\newcommand{\NholmfseedIbBEbp}{yes}
\newcommand{\NholmgseedIbDCbp}{yes}
\newcommand{\NholmfseedIbDCbp}{yes}
\newcommand{\Nholmm}{24}
\newcommand{\Nholmrawread}{15}
\newcommand{\Nholmglobalread}{7}
\newcommand{\Nholmfamilyread}{11}

\title{Context Injection and Long-Horizon Reliability: Separating Duplicate Recognition, Local Processing, and Counter Tracking}
\author{Thomas Serval \\ Baracoda AI Labs; NEOMA Business School}

\begin{document}
\maketitle

\begin{abstract}
Long-horizon failures of language models are usually measured as end-task success or as a horizon length. We instead score, step by step, three components of a 400-step ledger task in which every step is easy in isolation and the model's own answers are re-injected into its context: recognising a record already seen 5 to 300 steps earlier, processing the current record locally, and writing a counter incremented at every step. On Qwen3-8B (non-thinking, 20 seeds, replicated on a second machine and backend), with every step scored component by component, two results hold. First, duplicate-recognition errors rise late in the episode --- from \NmBrCbefore{} to \NmBrCafter{} per step at risk across step 200 --- while the counter makes no error in \NmBctrsteps{} steps and local processing is descriptively stable; the rise combines long-distance recall errors and rising false alarms. Second, a block of irrelevant content inserted before each step makes the counter fail in \NmCarmdctrseedshit{} of \NmCarmdseeds{} seeds on each machine, against none without it, while its effect on local processing is much smaller and not distinguishable from none. Three pre-registered follow-ups probe the counter. Its value depends on the value present in the response history: a corrupted value persists without correction in every seed, and an exact external reminder, repeated before each message, re-aligns it within one to two steps. A block with the exact form of the helpful one but irrelevant content was not conclusive. Under the irrelevant block, no counter error was observed when an exact compact reminder of the state, supplied externally by an oracle, preceded each message (\NpBAhitds{} of \NpBAseeds{} seeds with an error, against \NpBAhitd{} without it), and this reminder outperformed a length-matched neutral text; whether a single reminder suffices was not conclusive, and the absence of damage to the other components was not established. Pre-specified and complementary analyses are distinguished; family-wise results are reported with Holm correction (§4). An aggregate or exclusive error classification can hide the difference we report.
\end{abstract}

\section{Introduction}

A model that runs for hundreds of steps can fail because it can no longer connect what it sees now to what it saw long ago, because it mis-processes the current input, or because it loses track of a value it has to carry. End-task success and horizon length mix the three. Recent work shows that per-step accuracy decays with the number of turns and that errors in the history make later errors more likely \citep{sinha2025}; that context length alone degrades performance even when retrieval is perfect \citep{du2025,levy2024,hong2025}; and that agents drift from their goal under pressure \citep{arike2025}. What these results do not tell a practitioner is \emph{which} component gives way first, and whether adding context to help the model is safe.

We build a bench in which three step-level components are scored separately, and we read them as a \textbf{hazard}: the probability of an error at step \emph{k} given no earlier error of the same component. By \emph{carried state} we mean, operationally, a value the model must rewrite at every step from its own previous output --- here, a record counter; we make no claim about how the model represents it. By \emph{duplicate recognition} we mean answering \texttt{DOUBLON} exactly when the current code was seen earlier in the episode. The protocol does not measure intent or goals.

\textbf{Main results (Qwen3-8B).}

\begin{itemize}
  \item \textbf{R-A. Duplicate-recognition errors rise with the horizon; the counter makes none; local processing is descriptively stable.} The rise is carried by long recall distances and by false alarms on new records, not by a loss at constant distance.
  \item \textbf{R-B. Injected context disrupts the counter far more than local processing.} A block of irrelevant content before each step makes the counter fail in every seed; its effect on local processing is much smaller and not distinguishable from none, under a component-wise scoring that removes any masking between rules. Prior work established that length hurts; the observation here is \emph{which} component the tested blocks hit. A pre-registered follow-up (§5.3) shows that the counter depends on the value present in the response history; in another (§5.5), under the same repeated disruption, no counter error was observed when an exact reminder of the state, supplied by an oracle, preceded each message, and that reminder outperformed text of the same length without the state.
\end{itemize}

\textbf{Secondary results}, each reported with the multiple-comparison correction it survives: listing the already-seen codes before each step lowers duplicate-recognition errors, partially; repeating the rules before each step was followed by duplicate-recognition errors in every seed within 25 steps; over Qwen3 4B--32B, the duplicate-recognition hazard falls from 8B to 14B and to 32B; and the 32B tends to fail all-or-nothing.

\section{Related work}

\textbf{Long-horizon execution.} \citet{sinha2025} isolate execution from planning with a dictionary lookup and a running sum, on the same Qwen3 4B--32B ladder, and show that per-step accuracy decays, that errors in the history cause further errors (self-conditioning) independently of scale, and that thinking mode removes self-conditioning; in conversation, models that take a wrong turn do not recover \citep{laban2025}. We differ in three ways: we score components instead of turns; we vary the distance of the record to be recognised; and we test injected blocks against an irrelevant-content control. Sinha et al. also decompose their task in an appendix (lookup alone, addition alone, state tracking); what we add is the measurement of several components jointly, within the same trajectories, under the same interventions. Their running sum fails where our counter holds, and there is no contradiction: their sum requires a lookup plus a two-digit addition at every turn, our counter only a +1. That our counter holds without injection is consistent with plain continuation of an autoregressive pattern, which is why we call it \emph{trivial} and draw no conclusion from its stability alone. \citet{pai2026} carry an explicit state through 196 dependent tool calls and separate state errors from arithmetic errors; their \emph{implicit} round index slips, our \emph{explicit, written} counter does not. METR's time horizon \citep{kwa2025} is a whole-task, human-time measure; ours is a per-step hazard on trivial steps.

\textbf{Context length and distractors.} Performance degrades with position \citep{liu2023}, with input length on a fixed task, Qwen3-8B included \citep{levy2024,hong2025}, with irrelevant sentences \citep{shi2023}, across synthetic long-context suites \citep{hsieh2024}, with narrative wrapping \citep{pan2025}, and with length alone, even as whitespace or masked tokens \citep{du2025}. We do not claim that volume hurts. We claim that, on a counter the model writes and reads back at every step, the tested blocks hurt the counter far more than local processing. The "retrieve then solve" strategy of \citet{du2025} is close to listing the seen codes; neither they nor the other works compare it with an irrelevant block at least as long.

\textbf{Recognition, interference and drift.} The late rise of false alarms we observe --- new codes reported as already seen --- is compatible with proactive interference \citep{wang2025} as seen codes accumulate. \citet{arjmandi2026} uses "binding drift" and "commitment drift" for an agent whose state is owned by code, so that binding error is zero by construction; our duplicate recognition is defined operationally, and our rule re-injection places the rules in the prompt rather than in a code-owned store --- different objects, not contradictory results. Goal drift under competing objectives \citep{arike2025} motivates the question but is measured per episode.

\section{Bench}

\textbf{Task.} The model receives, once and at step 0, the rules of a ledger, then one record per message (a 4-character code, two integers, and a random token). For each record it answers one line with four fields: a counter (records processed so far, this one included), the code in upper case, the sum of the two integers, and the token copied verbatim. \textbf{R1 (local processing)}: code and sum must be right (the token is scored separately, as the canary). \textbf{R2 (duplicate recognition)}: if the code was already seen earlier in the batch, the model must answer \texttt{DOUBLON} instead; each episode contains repeats at distances 5, 40, 150 and 300 steps, placed by design so that a repeat at distance \emph{d} cannot occur before step \emph{d}. \textbf{Counter}: the counter must equal the step index + 1; in the original arms nobody ever reminds the model of it (the follow-up experiments of §5.3 and §5.5 add such reminders). \textbf{R3}: answer \texttt{FIN <total>} only after an end-of-batch message (part of the bench, not analysed here). Rules and records are in French.

\textbf{Self-conditioning is kept.} The model's \emph{own} previous answers are re-injected into its history, never the expected ones.

\textbf{Scoring.} Every step is scored \textbf{component by component} from the logged expected and observed lines (\texttt{harnais/renoter.py}): format, R2, counter, code, sum and token are each scored as correct, wrong, not applicable, or not observable, with no priority; a \texttt{DOUBLON} answer never counts as a local-processing success, and a step enters the risk set of a component only if that component is observable. The counter is observable on every readable line, including a \texttt{DOUBLON} answer, which still begins with the counter (\texttt{SORTIE: N DOUBLON code}, with N the counter); its denominator is therefore every step of every seed (\NmBctrsteps{} on M1 in Phase 0), whereas R1 is applicable only on computation steps. \textbf{This is the only scoring used for every number, table and figure of this paper.} The bench's original evaluator gave each line a single, exclusive label (order R3, format, R2, counter, R1), so that an error of higher rank hid an error of lower rank on the same line --- exactly the pattern that could manufacture R-B, since once the counter is off every later line was labelled "counter". It was replaced after an external review; all readings, including the sealed ones, are recomputed with the component-wise scoring, and the values the exclusive labels had given are listed in an audit appendix (Appendix C) as replaced, not as results. We document a correction; we do not keep a biased measure as a result. R2 is identical under both scorings; R1 in the injected arms is not.

\textbf{Validity controls.} A canary (the token to copy) detects a failure to copy the current input --- a collapse of reading: a run whose 20-step echo rate falls under 0.9 is excluded. It does not detect the loss of older history; truncation is controlled separately, by checking that no step exceeded the configured window. In the original study (Phase 0 and Phase 1 arms on both machines, and the confirmatory size map), none of the \Nvalepisodes{} episodes was excluded, and no step exceeded the window (at most \Nvalmaxtokens{} of \Nvalnumctx{} tokens). The follow-up campaigns are counted separately: experiments R and L, \NpIvalepisodes{} episodes, \NpIvalexcluded{} excluded; experiment S, \NpBAvalepisodes{} episodes, \NpBAvalexcluded{} excluded, plus \NpBAinvalidsleep{} episodes discarded and re-run because the machine slept (§5.5).

\textbf{Models and decoding.} Qwen3 via ollama, library tags, Q4\_K\_M quantisation; temperature 0, top-k 1, seed 0; context 40,960 tokens; non-thinking mode obtained by sending a raw ChatML prompt with a pre-closed think block (the 4B is the Instruct-2507 variant, which has none). M1 is an Apple-silicon Mac (ollama 0.32.5), M2 an NVIDIA A40 (ollama 0.32.11); \texttt{qwen3:8b} has the same digest on both. Everything ran locally; no remote API was used.

\textbf{Injected blocks (arms).} Arm 0 is bare. \textbf{Arm A (rules re-injected)} repeats the rules verbatim before every step. \textbf{Arm B (seen codes listed)} lists the codes already seen before the current step --- no new information, and never the verdict. \textbf{Arm D (irrelevant control)} injects operational prose with as many words as the rules plus as many fake codes as arm B's list: it is matched to the combination A+B (arm C, never run), hence longer than B. Median prompt size at step 100 on M2: \NmCarmAtokenssBAA{} tokens (arm 0), \NmCarmbtokenssBAA{} (B), \NmCarmdtokenssBAA{} (D). This imbalance cuts both ways. For the counter, the shorter block, B, damaged it more (§5.2); experiment L (§5.4) gives no evidence that relevance explains this. For duplicate recognition it works \emph{in favour of} B: if length alone hurt recognition, the longer D would be disadvantaged, so the B--D comparison on R2 overstates B's advantage by an unknown amount. Each block decorates the current message only; the bare message enters the history.

\section{Measurement}

\textbf{Hazard, not survival.} The index of the first failure is dominated by isolated one-letter errors, and a constant per-step error rate produces the same decaying survival curve as a genuine rise. We read the \textbf{hazard per component} in six step slices (0--25, 25--50, 50--100, 100--200, 200--300, 300--400); the sealed test compares steps before and after step 200 with a one-sided Fisher test, read at p < 0.01. Because errors within a seed are not independent, every main effect is also estimated at the seed level, with a bootstrap over seeds (10,000 resamples) that keeps seeds paired when the same seeds are used in two arms.

\textbf{Absence of an effect is not shown by a non-significant test.} Where the text says that a component is stable, it gives the interval of the effect, and the reader should read the upper bound.

\textbf{Multiple comparisons.} Three scopes are declared. (i) The \Nholmm{} p-values of the original study's confirmatory readings (Appendix B), recomputed with the component-wise scoring, are corrected with Holm's procedure at 0.01, in two ways. The \emph{reference} correction treats them as one family: \Nholmglobalread{} of the \Nholmrawread{} readings significant before correction survive it. A \emph{secondary} correction treats each claim as its own family (\Nholmfamilyread{} survive); these families were declared after the reference correction had been computed, so they never replace it. (ii) The two follow-up experiments of §5.3--5.4 have their own pre-registered family of two tests, Holm at 0.01, and experiment S (§5.5) has another (T1 and T2), Holm at 0.01. (iii) The robustness analyses added after the external reviews (decomposition of R2, seed-level intervals, interaction tests, restricted analyses) are reported with intervals; their few p-values are marked as secondary and stay outside any family.

\textbf{Noise floor.} With the model unloaded before each episode, \Nfloorunloadseeds{} seeds run \Nfloorunloadpasses{} times (\Nfloorunloadsteps{} steps) give a range of \Nfloorunloadrangefaults{} on the number of errors. The runs of this paper were chained without unloading; there, \Nfloorhotseeds{} seeds run four times (\Nfloorhotsteps{} steps) differ by at most \Nfloorhotrangefaults{} error, a hazard of \Nfloorhothazard{}.

\textbf{Size map.} For each size, the cutoff is the median first-failure step of a separate exploration set of 10 seeds; 20 confirmatory seeds follow. The sealed rule says "the median"; the analysis implemented its integer part (medians \NexplmedianEb{}, \NexplmedianIb{}, \NexplmedianBEb{} and \NexplmedianDCb{} for 4B, 8B, 14B and 32B); using the next integer step changes \Ncutoffroundingchanges{} of the sealed verdicts. Between sizes, comparisons are made both at each size's own cutoff and at a common cutoff (the highest own cutoff, step \Ncutcommon{}), with censoring treated as time at risk.

\section{Results}

\subsection{R-A: duplicate recognition fails late, the counter never does}

On M1, the duplicate-recognition (R2) hazard is \NmBrCbefore{} before step 200 (\NmBrCfbefore{} errors / \NmBrCnbefore{} steps at risk) and \NmBrCafter{} after it (\NmBrCfafter{} / \NmBrCnafter{}); one-sided Fisher p = \NmBrCp{}. On M2: \NmCrCbefore{} $\rightarrow$ \NmCrCafter{}, p = \NmCrCp{}. Both survive the reference correction, and the component-wise re-scoring leaves them unchanged. At the seed level, counting every R2 error, the rate after step 200 is \NrCtimemB{} the rate before it on M1 (95\% interval \NrCtimemBlo{}--\NrCtimemBhi{}) and \NrCtimemC{} on M2 (\NrCtimemClo{}--\NrCtimemChi{}). The counter makes no error: \NrenmBarmActrf{} in \NrenmBarmActrn{} steps on M1, \NrenmCarmActrf{} in \NrenmCarmActrn{} on M2. Local processing is descriptively stable --- between \NmBrBslicemin{} and \NmBrBslicemax{} per step in all six slices on M1 --- but not shown to be flat: its rate after step 200 is \NrBtimemB{} the rate before it on M1 (interval \NrBtimemBlo{}--\NrBtimemBhi{}) and \NrBtimemC{} on M2 (\NrBtimemClo{}--\NrBtimemChi{}), which does not exclude an increase of that size.

\textbf{What rises is not recognition at constant difficulty.} By design, repeats at distance 150 and 300 occur only late. At distances 5 and 40, the only ones present on both sides of step 200, misses do not rise: \NshortmBfbefore{}/\NshortmBnbefore{} before and \NshortmBfafter{}/\NshortmBnafter{} after on M1 (\NshortmCfbefore{}/\NshortmCnbefore{} and \NshortmCfafter{}/\NshortmCnafter{} on M2). The late errors are of two kinds: misses at long distance (\NdBFAmBfafter{}/\NdBFAmBnafter{} at distance 150 and \NdDAAmBfafter{}/\NdDAAmBnafter{} at 300 after step 200, M1), and \textbf{false alarms}, new codes reported as already seen: \NfamBfbefore{} in \NfamBnbefore{} new-code steps before step 200 against \NfamBfafter{} in \NfamBnafter{} after it on M1 (\NfamCfbefore{} and \NfamCfafter{} on M2). The rise combines long-distance recall errors and rising false alarms; the design does not separate the effects of time, context length and number of stored items.

\textbf{How the Phase 0 test was set.} The before/after-200 hazard test was developed on preliminary data: at the time the criterion was committed, only noise-floor replays of seeds 1--3 had been run and read. Those seeds are part of the analysed sample. On the remaining seeds 4--20 of M1 alone, the R2 hazard is \NpAhbefore{} before step 200 (\NpAhfbefore{}/\NpAhnbefore{}) and \NpAhafter{} after it (\NpAhfafter{}/\NpAhnafter{}; p = \NpAhpr{}, secondary), and the M2 replication was run after the criterion was fixed. We therefore describe this test as developed on preliminary data, evaluated on the remaining seeds and replicated on M2 --- not as pre-registered.

\begin{figure}[t]
\centering
\includegraphics[width=\linewidth]{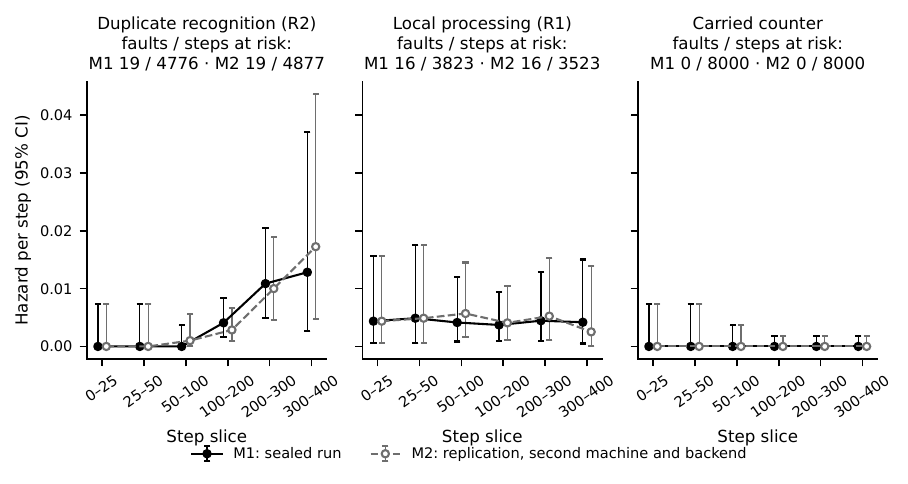}
\caption{\textbf{Duplicate recognition, local processing and the counter on qwen3:8b.} Per-step hazard (probability of an error at step k given no earlier error of the same component) by step slice, for duplicate recognition (R2), local processing (R1, code and sum) and the counter; component-wise scoring. 20 seeds per machine. Error bars: 95\% Clopper-Pearson. The R2 rise is decomposed in §5.1 (long distances and false alarms); R1 is descriptively stable; the counter makes no error.}
\label{fig:fig1_dissociation}
\end{figure}

On the second Phase 0 substrate (Qwen3-4B-Instruct, M1), the counter does fail (\NbEctrfaults{} errors in \NbEctrsteps{} steps at risk) and the R2 hazard is \NbErCbefore{} $\rightarrow$ \NbErCafter{} across step 200, p = \NbErCp{}: not read, with only \NbErCnafter{} steps at risk after the cutoff.

\subsection{R-B: injected context disrupts the counter far more than local processing}

\textbf{The counter.} This is a count. With no counter error in the bare arm, arm D makes the counter fail in \NmCarmdctrseedshit{}/\NmCarmdseeds{} seeds on M2 and \NmBarmdctrseedshit{}/\NmBarmdseeds{} on M1 (95\% interval for the proportion of seeds: \NseedscidmClo{}--1, against 0--\NseedsciAmChid{} in the bare arm; the same seeds are used in both arms, exact paired test p = \NmcnemardmCpr{}, secondary). Arm B does so in \NmCarmbctrseedshit{}/\NmCarmbseeds{} seeds. Once broken, the counter rarely recovers: under D, \NrenmCarmdctrf{} of \NrenmCarmdctrn{} lines carry a wrong counter on M2.

\textbf{Local processing under the same blocks}: the R1 error rate is \NrenmCarmArB{} per applicable step in the bare arm, \NrenmCarmdrB{} under D and \NrenmCarmbrB{} under B on M2. The paired seed-level ratio D/bare is \NrenmCarmdrBratio{} (95\% interval \NrenmCarmdrBlo{}--\NrenmCarmdrBhi{}) on M2 and \NrenmBarmdrBratio{} (\NrenmBarmdrBlo{}--\NrenmBarmdrBhi{}) on M1; B/bare is \NrenmCarmbrBratio{} (\NrenmCarmbrBlo{}--\NrenmCarmbrBhi{}) and \NrenmBarmbrBratio{} (\NrenmBarmbrBlo{}--\NrenmBarmbrBhi{}). These intervals do not show that local processing is unaffected --- on M1 the D/bare ratio is compatible with an increase of up to \NrenmBarmdrBhipct{}\% --- and we do not claim that. What they show is a difference of magnitude: the effect on the counter is far larger than any effect on local processing. That difference is estimated directly, per seed, as the effect on the counter minus the effect on R1 (both in errors per step): under D it is \NinterdmC{} (interval \NinterdmClo{}--\NinterdmChi{}) on M2 and \NinterdmB{} (\NinterdmBlo{}--\NinterdmBhi{}) on M1, positive in every seed (Wilcoxon p = \NinterdmCpr{}, secondary). A threshold on the R1 ratio (upper bound below 2) was used as the decision rule of the re-scoring (Appendix C); it was fixed after the original data had been seen, though before the re-scoring was run.

\emph{Denominators.} R1 is observable only on computation steps where the model did not answer \texttt{DOUBLON} and the line was parseable. Unobservable steps are rare: \NunobsmCarmApct{}\% of computation steps in the bare arm, \NunobsmCarmbpct{}\% under B and \NunobsmCarmdpct{}\% under D on M2 (all from a \texttt{DOUBLON} answer, none from a broken format). Restricting the comparison to steps where R1 is observable in all three arms for the same seed (\NcmpmCsteps{} steps on M2) gives D/bare \NcmpmCdratio{} (\NcmpmCdlo{}--\NcmpmCdhi{}) and B/bare \NcmpmCbratio{} (\NcmpmCblo{}--\NcmpmCbhi{}); on M1, D/bare \NcmpmBdratio{} (\NcmpmBdlo{}--\NcmpmBdhi{}) and B/bare \NcmpmBbratio{} (\NcmpmBblo{}--\NcmpmBbhi{}).

\textbf{B damages the counter more than D}, although D is longer: by \NmCbdctrratio{} on M2 (p = \NmCbdctrp{}) and \NmBbdctrratio{} on M1 (p = \NmBbdctrp{}); this survives the per-claim correction but not the reference one.

\begin{figure}[t]
\centering
\includegraphics[width=\linewidth]{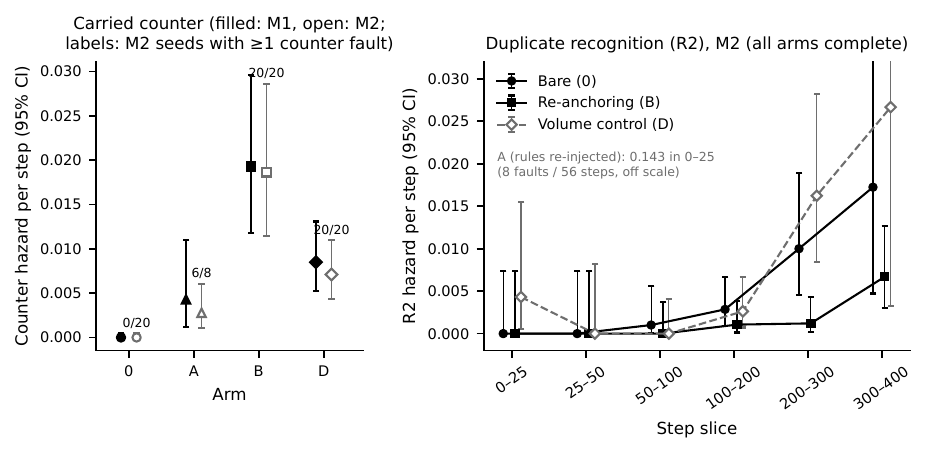}
\caption{\textbf{Injected blocks (arms 0/A/B/D).} Left: counter hazard per arm, both machines; labels give the number of M2 seeds with at least one counter error. Right: R2 hazard by slice on M2, where all four arms are complete. Arm A (rules re-injected) makes an R2 error in the first slice in every seed and leaves the risk set; it is reported as text, off scale. B = already-seen codes listed before each step; D = irrelevant-content control, matched to rules + codes and therefore longer than B (not an equal-token control; see §3).}
\label{fig:fig2_interventions}
\end{figure}

\begin{table}[t]
\centering
\small
\resizebox{\linewidth}{!}{%
\begin{tabular}{lllllll}
\toprule
arm & seeds & counter hazard & seeds with a counter error & R2 hazard & R1 error rate & prompt tokens at step 100 \\
\midrule
0 bare & \NmCarmAseeds{} & \NmCarmActr{} & \NmCarmActrseedshit{} & \NmCarmArC{} & \NrenmCarmArB{} & \NmCarmAtokenssBAA{} \\
A rules re-injected & \NmCarmaseeds{} & \NmCarmactr{} & \NmCarmactrseedshit{} & \NmCarmarC{} & \NrenmCarmarB{} & --- \\
B seen codes listed & \NmCarmbseeds{} & \NmCarmbctr{} & \NmCarmbctrseedshit{} & \NmCarmbrC{} & \NrenmCarmbrB{} & \NmCarmbtokenssBAA{} \\
D irrelevant control & \NmCarmdseeds{} & \NmCarmdctr{} & \NmCarmdctrseedshit{} & \NmCarmdrC{} & \NrenmCarmdrB{} & \NmCarmdtokenssBAA{} \\
\bottomrule
\end{tabular}}
\caption{Arms on M2 (primary; M1 replication in the text), component-wise scoring. Hazards are censored at the first error of the component, over steps where it is observable; the R1 error rate uses all applicable steps.}
\label{tab:arms}
\end{table}

\subsection{Pre-registered follow-up: what does the counter depend on? (experiment R)}

To ask whether the counter depends on what the model has written before, or is re-derived independently of it, we ran a follow-up after the external review, sealed before any measurement (Appendix A) on machine M1, after a bridge check that reproduced the August error counts exactly (largest difference \NpIpontmaxecart{} error per seed). At the first computation step from step 100 on, the model's own line entered its history with its counter replaced by the true value plus 10; everything else was unchanged, and the run continued to step 200 (\NpIrAseeds{} seeds). In arm R0 nothing else was added; in arm RR \emph{each} later message was preceded by a compact reminder of the true state ("last valid counter = N") --- the reminder was repeated at every step, so we do not show that a single correction would suffice. Each seed was classified on the 5 steps 16--20 steps after the injection as \emph{re-aligned} (true counter), \emph{propagated} (true + 10) or other.

Without the reminder, \textbf{\NpIrApropage{} of \NpIrAseeds{} seeds propagated the injected offset} and none re-aligned (95\% interval for the proportion propagated: \NpIrAicprolo{}--\NpIrAicprohi{}): the sealed verdict is \emph{\NpIrAverdict{}}, the prediction we had signed --- a name that over-interprets the measured behaviour, as explained below. With the reminder, \NpIrrrecale{} of \NpIrrseeds{} seeds re-aligned, after a median of \NpIrrdelaimedian{} step (at most \NpIrrdelaimax{}). The same seeds were used in both arms, so the conforming analysis is a paired one: \NpIpairsdisc{} of \NpIpairsdisc{} discordant pairs favour RR and \NpIpairsrev{} the reverse (exact paired test, one-sided, p = \NpItestBpairedpr{}). The sealed test was a Fisher test for independent groups (p = \NpItestBpr{}), a design error of the pre-registration that we report rather than hide; both survive Holm at 0.01 over the follow-up's two tests.

The result is behavioural: \textbf{the counter depends on the value present in the response history; a corrupted value persists without correction, and an exact external reminder re-aligns it within one to two steps.} Several mechanisms produce this behaviour --- treating the previous value as authoritative, applying a "previous + 1" rule, or continuing a pattern --- and the task itself asked the model to carry the previous value forward. The reminder is supplied by an experimental oracle: arm RR shows that the model \emph{uses} an explicit state when one is provided, not that it tracks the counter internally.

\subsection{Pre-registered follow-up: content or form? (experiment L)}

In the original arms, the irrelevant control D differed from B in more than relevance: its header, its prose and its length. Experiment L removes every difference but one. Its block has B's header, separators, position, number of codes and code template, with codes that never occur in the ledger; the prompt sizes of L and B differ by a median of \NpIltokdiffpct{}\% at the same seed and step. B was re-run alongside L on M1 (\NpIlseeds{} paired seeds, 200 steps), because the August B runs on M1 had shared the inference server.

The sealed criterion compared the counter hazards of L and B with an equivalence margin of 0.8--1.25. The ratio L/B is \NpIlratio{} (95\% interval \NpIllo{}--\NpIlhi{}): the verdict is \emph{\NpIlverdict{}} --- neither a difference nor an equivalence is shown --- and the second test of the follow-up's family (paired Wilcoxon on the step of the first counter error, p = \NpItestCpr{}) \NpIholmtestC{} Holm. Our signed prediction, that relevance matters, is \textbf{not supported}. Descriptively, L behaves like B, not like D: the counter fails in \NpIlhitl{} of \NpIlseeds{} seeds under L and \NpIlhitb{} under B (none in the bare arm over the same 200 steps), with a median first error at step \NpIlfirstl{} under L and \NpIlfirstb{} under B. L gives no evidence that relevance explains the B--D difference; form remains a hypothesis this experiment could not confirm. Local processing under L, against the bare arm, gives a ratio of \NpIlrBratio{} (\NpIlrBlo{}--\NpIlrBhi{}). False alarms remain rare under both (\NpIlfal{} under L and \NpIlfab{} under B, in \NpIlfan{} new-code steps).

\subsection{Pre-registered follow-up: repair under repeated disruption (experiment S)}

Experiment R showed that an exact reminder, repeated at every step, re-aligns a counter corrupted \emph{once}. Experiment S (for \emph{state}) asks the harder question, fixed before the runs as the last experiment of this paper: under a disruption repeated at every step (the irrelevant block D before each message), does an exact, compact reminder of the state repair the counter; is it the state information itself that repairs, rather than any added text of the same length; and does a single correction persist? It was sealed before any measurement, with an amendment sealed before launch (Appendix A), on machine M1 (same model, digest and decoding as in August).

\textbf{Conditions.} On the same \NpBAseeds{} seeds and 200 steps: D (the original block before each message); DS (D, then "last valid counter = N" just before the current message, at every step from the second one); D1 (the same reminder \emph{once}, at the step after the seed's first counter error, then nothing); DN (D, then a neutral text --- "no tracking information" followed by dots --- at the same position, whose length in tokens was calibrated before the first episode to equal the reminder's for the same number of digits). N is exact by construction: the number of messages already received, computed from past inputs only. The reminder is an experimental oracle, as in experiment R. The bare condition was not re-run: by amendment, the August bare episodes of M1, cut to their first 200 steps, serve as a descriptive reference only.

\textbf{Sealed criteria.} Repair: the ratio DS/D of the proportion of lines with a wrong counter (pooled over seeds), 95\% paired bootstrap interval by seed; \emph{repairs} if the upper bound is at most 0.2 (a reduction of at least 80\%), \emph{does not repair} if the lower bound exceeds 0.2, otherwise not conclusive. No damage: the same ratios for local processing (R1) and duplicate recognition (R2); \emph{no damage} if both upper bounds are below 1.5. Persistence under D1: among seeds whose first counter error falls before step 199, the proportion that re-align for 5 steps after the reminder and then fail again, Clopper--Pearson interval; \emph{relapse} if the lower bound exceeds 0.5, \emph{persistence} if the upper bound is below 0.5. The follow-up's own family has two paired Wilcoxon tests, Holm at 0.01: T1, the wrong-counter proportion is lower under DS than under DN (one-sided); T2, DN against D (two-sided). Our signed bets were: DS repairs, DN does not, a single reminder relapses.

\begin{figure}[t]
\centering
\includegraphics[width=0.5\linewidth]{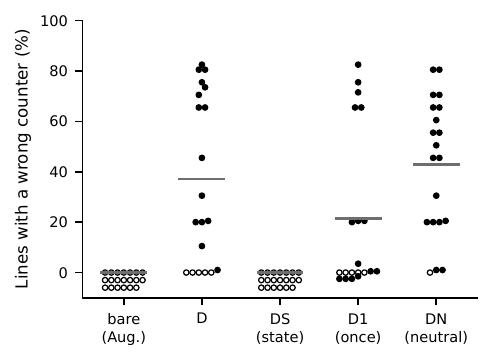}
\caption{\textbf{Experiment S: repair under repeated disruption.} Proportion of lines with a wrong counter, one dot per seed (20 seeds, 200 steps, same seeds in every condition); open dots: no counter error; bar: proportion pooled over seeds. D: irrelevant block before each message; DS: D plus an exact reminder of the state ("last valid counter = N", supplied by an oracle) before each message; D1: the same reminder once, after the seed's first counter error; DN: D plus neutral text of the reminder's length in tokens. Bare: the August bare runs of the same machine, first 200 steps (descriptive reference, not re-run).}
\label{fig:fig5_state_reminder}
\end{figure}

\textbf{Results.} The counter failed in \NpBAhitd{} of \NpBAseeds{} seeds under D, \NpBAhitds{} under DS, \NpBAhitdn{} under DN and \NpBAhitdone{} under D1 (\NpBAhitbare{} in the August bare runs over the same 200 steps); the proportion of lines with a wrong counter was \NpBApropd{}\% under D, \NpBApropds{}\% under DS and \NpBApropdn{}\% under DN (Figure \ref{fig:fig5_state_reminder}). The sealed ratio DS/D is \NpBArepratio{}, and its sealed bootstrap interval is \NpBAreplo{}--\NpBArephi{}: the sealed verdict is \emph{\NpBArepverdict{}}. That interval is \textbf{degenerate} --- DS has no error in any seed, so every resample gives the same ratio --- and it quantifies nothing; we report it because it was the sealed computation, and we do not use it to claim a reduction of at least 80\% in any wider population. Two complementary analyses, \textbf{added after review}: the proportion of seeds with a counter error under DS is \NpBAhitds{}/\NpBAseeds{} (95\% Clopper--Pearson interval \NpBAdscplo{}--\NpBAdscphi{}); and, the seeds being paired, \NpBApairdisc{} seeds failed under D and not under DS, \NpBApairrev{} the reverse (exact paired test, one-sided, p = \NpBApairpr{}). For local processing the ratio DS/D is \NpBAdmgrBratio{} (\NpBAdmgrBlo{}--\NpBAdmgrBhi{}) and for duplicate recognition \NpBAdmgrCratio{} (\NpBAdmgrClo{}--\NpBAdmgrChi{}): \emph{\NpBAdmgverdict{}}. T1 (DS below DN) has p = \NpBAtBpr{} and \NpBAholmtB{} Holm; T2 (DN against D) has p = \NpBAtCpr{} and \NpBAholmtC{} Holm. Under D1, \NpBAdoneana{} seeds were analysable (the other \NpBAdonenonana{} made no counter error at all); \NpBAdonerecale{} re-aligned for 5 steps and \NpBAdonerechute{} of them later relapsed. On the sealed denominator, the analysable seeds, the proportion of relapses is \NpBAdonerechute{}/\NpBAdoneana{} (95\% interval \NpBAdonelo{}--\NpBAdonehi{}): \emph{\NpBAdoneverdict{}}. This composite criterion cannot establish durable recovery, because episodes that never recover also do not satisfy the recovery-then-relapse event. Among the seeds that re-aligned it is \NpBAdonerechute{}/\NpBAdonerecale{} (\NpBArelreclo{}--\NpBArelrechi{}; added after review). Table \ref{tab:d1_seeds} lists every analysable seed.

\begin{table}[t]
\centering
\small
\resizebox{\linewidth}{!}{%
\begin{tabular}{lll}
\toprule
outcome & seeds & seed (first counter error) \\
\midrule
no recovery & \NpBAcatnonen{} & \NpBAcatnoneseeds{} \\
brief recovery & \NpBAcatbriefn{} & \NpBAcatbriefseeds{} \\
recovery held to step 200 & \NpBAcatdurablen{} & \NpBAcatdurableseeds{} \\
recovery, then relapse & \NpBAcatrelapsen{} & \NpBAcatrelapseseeds{} \\
too little follow-up & \NpBAcatlaten{} & \NpBAcatlateseeds{} \\
\bottomrule
\end{tabular}}
\caption{Experiment S, single reminder (D1): outcome of every analysable seed, with the step of its first counter error in parentheses; the reminder came at the next step. \emph{No recovery}: the counter was never true again. \emph{Brief recovery}: true at the next step, wrong again within 5 steps. \emph{Recovery held}: true for 5 steps and never wrong again. \emph{Recovery, then relapse}: true for 5 steps, wrong later. \emph{Too little follow-up}: fewer than 5 steps left after the reminder. Categories defined after review; descriptive.}
\label{tab:d1_seeds}
\end{table}

Against our signed bets. \textbf{No counter errors were observed under the repeated reminder in the tested episodes}, against \NpBAhitd{} seeds with an error under D. \textbf{The exact reminder tested outperforms this length-matched neutral control} (T1). T2 does not separate DN from D, which does not show that they are equivalent. This does not establish that the state information alone explains the effect: the wording of the reminder and its instructive character differ from the neutral text too. \textbf{Whether a single correction persists is not conclusive}; we had bet on relapse. Descriptively, the single reminder had no visible effect in \NpBAcatnonen{} of the \NpBAdoneana{} analysable seeds: the next counter continued from the model's own wrong value, and never returned to the true one. Where the counter became true again, it did so within \NpBAdonedelaimax{} step. The proportion of wrong-counter lines was \NpBApropdone{}\% under D1.

The no-damage criterion is not met: local processing errors are rare (\NpBArBd{}\% of applicable steps under D, \NpBArBds{}\% under DS), and the upper bound of the ratio, \NpBAdmgrBhi{}, exceeds the sealed margin of 1.5. The absence of degradation of the other components is therefore not established. Duplicate-recognition errors were lower under DS (\NpBArCds{}\% of steps, against \NpBArCd{}\% under D, \NpBArCdn{}\% under DN and \NpBArCbare{}\% in the August bare runs); this is a secondary observation, not a pre-registered criterion, and we do not interpret it. The reminder is an oracle: experiment S shows that the model \emph{uses} an exact state supplied at every step, even under a repeated disruption, not that a deployed system could supply one. The state came from an oracle and the effective condition repeated it at every step; the necessary frequency was not established.

\textbf{Cost and conduct.} Median prompt size at step 100 was \NpBAtokd{} tokens under D, \NpBAtokds{} under DS and \NpBAtokdn{} under DN; median episode time was \NpBAlatd{} min under D and \NpBAlatds{} min under DS. The Mac was sometimes unplugged or put to sleep during the campaign: as sealed in the amendment, every episode crossed by a sleep (\NpBAinvalidsleep{}) was discarded and re-run from its start with the same seed, never resumed mid-episode; power source was logged at the start and end of each episode; no valid episode ran on battery (\NpBAbatteryeps{}), so latency is reported on mains only. No result was read before the last episode of the last seed; the reading was then run once, as sealed.

\subsection{Secondary: listing the seen codes, repeating the rules}

\textbf{Listing the seen codes (arm B) lowers duplicate-recognition errors, partially.} On the sealed hazard (first error per seed), B is below D before correction on both machines (p = \NmBbdrCp{} on M1, \NmCbdrCp{} on M2); after correction this holds on M1 only (per-claim correction). Against the bare arm, B is below it before correction on M1 only (p = \NmBbArCp{}; M2 p = \NmCbArCp{}). Counting every R2 error at the seed level (secondary), B's rate is \NbArCmC{} that of the bare arm on M2 (interval \NbArCmClo{}--\NbArCmChi{}) and \NbdrCmC{} that of D (\NbdrCmClo{}--\NbdrCmChi{}), with the caveat of §3 on block length. The irrelevant block D, compared with the bare arm, gives a ratio of \NdArCmC{} (\NdArCmClo{}--\NdArCmChi{}) on M2: we cannot exclude that D itself doubles R2 errors, so the earlier statement that "D is not worse than the bare arm" is withdrawn. The pre-registered repair criterion --- no rise within arm B, p > 0.05 --- is \textbf{not met} (p = \NmCarmbrCrisep{} on M2, \NmBarmbrCrisep{} on M1). We also note that this criterion was badly designed: it rewards a lack of power, since fewer errors make "p > 0.05" easier to reach. It is a weakness of our pre-registration; we report it and do not replace it after the fact. The comparison of D with the bare arm on the sealed hazard (p = \NmCdArCpworse{}) is likewise uninformative about the absence of an effect.

\textbf{Repeating the rules (arm A).} Under A, every seed made a duplicate-recognition error within the first slice: \NmCarmarCfirstf{} errors in \NmCarmarCfirstn{} steps at risk (hazard \NmCarmarCfirst{}), against \NmCarmArCfirstf{} in the bare arm; the counter failed in \NmCarmactrseedshit{}/\NmCarmaseeds{} seeds. With \NmCarmaseeds{} seeds (M2) and \NmBarmaseeds{} (M1), this is an observation of a large effect, not a tested one.

\subsection{Secondary: size map}

\begin{figure}[t]
\centering
\includegraphics[width=\linewidth]{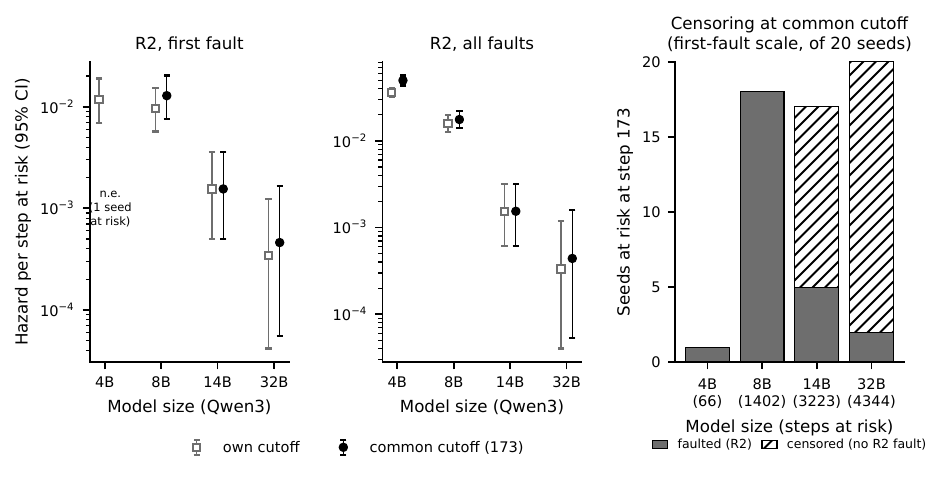}
\caption{\textbf{Size map, duplicate-recognition (R2) hazard in time at risk.} Left and centre: R2 hazard after each size's own cutoff (median survival of a separate exploration set, open squares) and after the common cutoff (highest own cutoff, filled circles), on the first-fault and all-faults scales. Error bars: exact Poisson 95\%. Log scale. Estimates resting on fewer than five seeds at risk are not plotted ("n.e."): at the common cutoff this is the 4B on the first-fault scale, which is used in no size comparison. Right: seeds still at risk at the common cutoff, split into those that later fault on R2 and those censored (no R2 fault by step 400); censoring grows with size. Seed-level tests are reported in the text before step-level tests.}
\label{fig:fig3_size_map}
\end{figure}

Figure \ref{fig:fig3_size_map} and Table \ref{tab:sizes} give, for Qwen3 4B, 8B, 14B and 32B, the R2 hazard at own and common cutoffs, on the first-error and all-errors scales, with seeds at risk, censored seeds and steps at risk. Qwen3-1.7B failed the admission barrier (exploration median \NcutBHb{}); an alternative scaffold for the 4B was invalidated by the canary in \NaltEbinvalid{}/\NaltEbtotal{} episodes. Re-scored component by component, the counter fails in \NrensizectrseedsEb{} of 20 seeds at 4B, in none at 8B or 14B, and in \NrensizectrseedsDCb{} at 32B; the sealed prediction was zero at 14B \emph{and} 32B, and it is refuted for the 32B.

\begin{table}[t]
\centering
\small
\resizebox{\linewidth}{!}{%
\begin{tabular}{lllllllll}
\toprule
size & own cutoff & hazard, own cutoff & seeds at risk (common) & censored & errors & steps at risk & hazard, common (bound) & hazard, all errors, common \\
\midrule
4B & \NcutEb{} & \NffownEbh{} & \NffcomEbarisque{} & \NffcomEbcensurees{} & \NffcomEbfautes{} & \NffcomEbpas{} & \NffcomEbh{} (\NffcomEbbound{}) & \NafcomEbh{} \\
8B & \NcutIb{} & \NffownIbh{} & \NffcomIbarisque{} & \NffcomIbcensurees{} & \NffcomIbfautes{} & \NffcomIbpas{} & \NffcomIbh{} (\NffcomIbbound{}) & \NafcomIbh{} \\
14B & \NcutBEb{} & \NffownBEbh{} & \NffcomBEbarisque{} & \NffcomBEbcensurees{} & \NffcomBEbfautes{} & \NffcomBEbpas{} & \NffcomBEbh{} (\NffcomBEbbound{}) & \NafcomBEbh{} \\
32B & \NcutDCb{} & \NffownDCbh{} & \NffcomDCbarisque{} & \NffcomDCbcensurees{} & \NffcomDCbfautes{} & \NffcomDCbpas{} & \NffcomDCbh{} (\NffcomDCbbound{}) & \NafcomDCbh{} \\
\bottomrule
\end{tabular}}
\caption{Size map, R2, time at risk. Own cutoff = integer part of the median first failure of each size's exploration set; common cutoff = step \Ncutcommon{}. "Seeds at risk" and "censored" are on the first-error scale (a seed that made an R2 error before the common cutoff is not at risk); the text's count of seeds \emph{without any} R2 error after the cutoff uses the all-errors scale over all 20 seeds, hence different numbers at 14B (\NffcomBEbcensurees{} of \NffcomBEbarisque{} against \NseednofaultBEb{} of 20). In brackets, the one-sided 95\% Poisson upper bound. The 4B at the common cutoff rests on one seed and enters no comparison.}
\label{tab:sizes}
\end{table}

\emph{Sealed within-size reading} (own cutoff): the 8B R2 hazard rises (\NsealedIbrCbefore{} $\rightarrow$ \NsealedIbrCafter{}, p = \NsealedIbrCp{}); the 14B and 32B are not conclusive (expected errors under the null \NsealedBEbrCexpectedhA{} and \NsealedDCbrCexpectedhA{}), not flat.

\emph{Between sizes}, at the common cutoff: per-seed hazards separate 8B from 14B (Mann-Whitney p = \NseedIbBEbp{}) and 8B from 32B (p = \NseedIbDCbp{}), but not 14B from 32B (p = \NseedBEbDCbp{}); the two separations, and the step-level tests, survive both corrections. On the first-error scale the ratios are \NffcomIbBEbratio{} (8B/14B, step-level p = \NffcomIbBEbp{}) and \NffcomIbDCbratio{} (8B/32B, p = \NffcomIbDCbp{}); at own cutoffs \NffownIbBEbratio{} and \NffownIbDCbratio{}. We do not claim a constant factor per size step: beyond 14B we lack the power to distinguish (the 32B has \NffcomDCbfautes{} R2 errors). Censoring is ordered by size --- \NseednofaultIb{}, \NseednofaultBEb{} and \NseednofaultDCb{} of 20 seeds without any R2 error after step \Ncutcommon{} (8B, 14B, 32B) --- and time at risk accounts for it only under the assumption that censoring is uninformative.

\subsection{Secondary: failure shape}

\begin{figure}[t]
\centering
\includegraphics[width=0.5\linewidth]{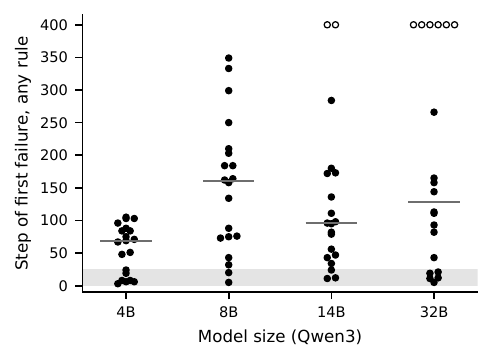}
\caption{\textbf{Failure shape by size.} Step of the first failure on ANY rule (not duplicate recognition alone), one dot per seed; open dots: no failure within 400 steps; bar: median; shaded band: failure within 25 steps.}
\label{fig:fig4_failure_shape}
\end{figure}

Figure \ref{fig:fig4_failure_shape} shows the step of the first failure \textbf{on any rule}, one dot per seed. \NextIb{}/20 8B seeds are at the extremes (failure within 25 steps, or none in 400), against \NextDCb{}/20 for the 32B (\NearlyDCb{} within 25 steps, \NfullDCb{} never; p = \NextIbDCbp{}, which survives the per-claim correction but not the reference one). The 14B is intermediate and not read (\NextBEb{}/20, p = \NextIbBEbp{}). If replicated, this pattern would motivate analysing early and later failures separately; it does not establish that 32B failures are predominantly immediate.

\subsection{Complementary evidence on self-conditioning (exploratory)}

The sealed R2 hazard is censored at the first error of the \emph{same} component, so earlier R1 errors can sit in the history, and self-conditioning \citep{sinha2025} could contribute to the late rise. We decided after seeing the data to censor at the first error of \emph{any} component, so that every counted step has an error-free history. The R2 hazard still rises across step 200: \NcleanmBfbefore{}/\NcleanmBnbefore{} $\rightarrow$ \NcleanmBfafter{}/\NcleanmBnafter{} on M1 and \NcleanmCfbefore{}/\NcleanmCnbefore{} $\rightarrow$ \NcleanmCfafter{}/\NcleanmCnafter{} on M2. With \NcleanmBfafter{} errors after the cutoff on each machine, this is complementary evidence, not a test.

\section{Discussion}

\textbf{What breaks when the counter breaks?} A trivial +1 counter may be carried by nothing more than the value in the model's own previous lines, and an injected block may break it by disrupting the layout of the prompt rather than by removing any stored value. Experiment R (§5.3) establishes the behaviour: the counter depends on the value present in the response history --- an injected wrong value is propagated by every seed, and a reminder repeated at every step re-aligns it within one to two steps. It does not tell which mechanism produces this. On what breaks the counter, a draft of this paper argued from the original arms that \emph{content} breaks the counter: B, the shorter block, broke it more and much earlier than D (first counter error at a median step of \NmCarmbctrfirstmedian{} under B against \NmCarmdctrfirstmedian{} under D on M2), most often by a missed increment (\NmCarmbctrfirstminusB{} of \NmCarmbctrseedshit{} seeds under B). Experiment L does not support that argument: a block with B's exact form and irrelevant codes breaks the counter about as early and as often as B (§5.4), although the pre-registered comparison is not conclusive. L gives no evidence that relevance explains the B--D difference; form remains a hypothesis this experiment could not confirm. We withdraw the content-based reading.

\textbf{What repairs the counter?} Under the irrelevant block repeated at every step, no counter error was observed when an exact statement of the state was placed just before each message, and that statement outperformed neutral text of the same length (§5.5). The added length alone does not explain the result; the state information, the wording of the reminder and its instructive character remain confounded. A single statement, placed once after the first error, often lost against the model's own history --- in some seeds it had no visible effect at all --- which is what one expects if the counter follows the value present in the history (§5.3). Two limits matter for practice: the state came from an oracle, and the effective condition repeated it at every step; the necessary frequency was not established. A system that relies on this needs an external source of state it can trust, which this paper does not provide.

\textbf{Adding context has a price on the counter, for the blocks we tested.} Retrieval and memory layers add text before the current message. On this bench, each of the four blocks tested --- rules, seen codes, irrelevant prose, irrelevant codes in the form of the seen codes --- broke a counter that was otherwise error-free. The one block that stated the state exactly, the reminder of experiment S, did not (§5.5). A layer that helps duplicate recognition may therefore hurt carried state, and must be evaluated on both. We did not test blocks of other kinds, sizes or positions.

\textbf{Duplicate recognition fails late.} The late rise of R2 combines long-distance recall errors and false alarms that appear only once many codes have been seen; the false alarms are compatible with interference between stored items, and the design does not separate the effects of time, context length and number of stored items.

\textbf{Scale.} Larger models recognise duplicates better over the same region of the episode. They may also fail differently, but that reading survives only the per-claim correction, and should be confirmed before anyone plans a deployment around it.

\section{Conclusion}

This paper is a diagnosis followed by one pre-specified repair test. The diagnosis: on a long, easy, self-conditioned task, some context additions strongly disrupt counter tracking while their effect on local processing is much smaller, and an aggregate or exclusive error classification can hide this difference; duplicate recognition fails late, through long-distance misses and false alarms. The counter follows the value present in the model's own history: a corrupted value persists, and an explicit external statement of the state restores it. The repair test: under a disruption that otherwise breaks the counter, no counter error was observed when an exact statement of the state preceded each message, while the absence of degradation of the other components was not established by the sealed criterion; the statement outperformed text of the same length without the state, and a single statement did not reliably hold. This motivates a design hypothesis, not a finding --- that carried state which matters should be stated explicitly and often rather than left to the model's own history --- since we tested neither several frequencies nor several kinds of state; and it is not an autonomous system: the state here came from an oracle, and supplying it reliably is a separate problem. Scoring components independently made these differences visible; the original exclusive error classification had masked some local-processing errors.

\section{Limitations and what we do not claim}

\begin{itemize}
  \item One synthetic task family, in French, with a trivially incremented counter; the counter result must not be generalised to arithmetic state; \citet{sinha2025} find the opposite for a running sum.
  \item One model family, Q4\_K\_M via ollama, greedy, \textbf{non-thinking}; thinking mode may change the picture; \citet{sinha2025} find that it removes self-conditioning.
  \item 20 seeds per condition; arm A has \NmCarmaseeds{} (M2) and \NmBarmaseeds{} (M1) seeds.
  \item Arm D is matched to the combination A+B, hence longer than B; arm C was not run.
  \item Distance and position in the episode are confounded by design.
  \item Experiment R establishes a behaviour, not a mechanism; its reminder was repeated at every step and supplied by an oracle. Experiment L gives no evidence that relevance matters, and does not establish that form does: its comparison is not conclusive (§5.4), and our signed prediction was not supported.
  \item Experiment S (§5.5) ran on one machine, with an oracle state, a trivial counter and one kind of disruption; its no-damage criterion and its persistence criterion are not conclusive, and our bet on relapse is neither supported nor refuted. Its sealed repair interval is degenerate (§5.5); the complementary interval and paired test were added after review. The reminder was repeated at every step, and its necessary frequency was not established. The bare reference is the August bare run, not a concurrent one.
  \item The bench's original exclusive labels hid lower-ranked errors; every reading was recomputed with the component-wise scoring (Appendix C lists the replaced values).
  \item Where the text reports stability, it rests on intervals whose upper bounds do not exclude increases of up to about twofold (R1 over time; R1 under the injected blocks; R2 under D).
  \item The Phase 0 hazard test was developed on preliminary data (seeds 1--3) that are part of the analysed sample; it is not described as pre-registered (§5.1).
  \item The sealed test of experiment R assumed independent groups although seeds were paired; the paired test is reported alongside it.
  \item The pre-registered repair criterion for arm B (p > 0.05 within the arm) rewards low power; it was a poor criterion, reported as written.
  \item A sealed prediction was refuted: the size map predicted no counter error at 14B and 32B; the 32B made \NrensizectrerrDCb{}.
  \item The Phase 1 arms on M1 ran while another job shared the machine's inference server; M2 is the primary source for the arms.
  \item The model digests of the 14B and 32B were not recorded before the rented machine was deleted.
  \item No pre-registration of the original study declared a family of tests; the common-cutoff comparison, the clean-history check and every robustness analysis of the external-review rounds were decided after seeing the data. The R1 decision threshold of Appendix C was fixed after the original data had been seen, before the re-scoring.
  \item \textbf{We do not claim} that a small model with external state beats a large model without it (not tested here); that the protocol measures intent or goals; that listing seen codes \emph{repairs} duplicate recognition; that each size step divides the hazard by a constant factor; that errors disappear beyond 8B (14B and 32B are not conclusive within size, not flat); or that context volume degrading performance is new.
\end{itemize}

\section{Reproducibility and AI assistance}

Code, all logged episodes, pre-registrations and seals are available at \texttt{https://github.com/Baracoda-ai-labs/context-injection-long-horizon} (MIT for code, CC BY 4.0 for data). This paper, its PDF and its \texttt{paper/numbers.json} correspond to the tag \texttt{v1.0-arxiv} of that repository. To reproduce it: \texttt{git clone} the repository, \texttt{git checkout v1.0-arxiv}, then \texttt{bash reproduce.sh}, which regenerates every number, table and figure from the logs without calling a model (Python 3 with scipy and matplotlib); it also runs \texttt{verify\_seals.py}, which checks the seals, and \texttt{harnais/verifier\_papier.py}, which re-derives the key numbers with independent code. \texttt{python3 harnais/renoter.py} re-scores every step component by component. The repository was checked to reproduce identically from a fresh clone in a clean directory. Experiments were run and analysed, and this draft was written, with the assistance of Claude Code and Claude (Anthropic); all design decisions and every sealed pre-registration were reviewed by the author.

\bibliography{refs}
\IfFileExists{tmlr.bst}{\bibliographystyle{tmlr}}{\bibliographystyle{plainnat}}

\appendix
\section{Pre-registration, seals and measurement rules}

\textbf{What the seals prove.} Every confirmatory campaign has a pre-registration whose SHA-256 hash is recorded (eleven seals, all verified by the package; the ninth covers the follow-up experiments of §5.3--5.4 and the last two experiment S (§5.5), each committed to the repository before any of their episodes was run). A hash proves that a document has not changed since it was hashed; it does not prove when. Anteriority rests on the repository's git history (commit dates are set by the author's machine; the repository is mirrored on GitHub). By that history: the Phase 1 pre-registration was committed at \NdatephaseBprereg{}, before the first Phase 1 data (\NdatephaseBdata{}); arm A at \Ndatearmaprereg{}, before its data (\Ndatearmadata{}); the size map at \Ndatesizemapprereg{}, before its data (\Ndatesizemapdata{}). For Phase 0, the first pre-registration (\NdatephaseAprereg{}) specified a survival criterion that was later superseded; the hazard criterion actually used was committed at \NdatephaseAhazardcriterion{}, before the 20-seed data were complete, after three noise-floor seeds had suggested it, and its formal seal (\NdatephaseAhazardseal{}) postdates the reading of the hazard table. At the time of that commit the repository held only the noise-floor replays of seeds 1--3; seeds 4--20 had not been run. The Phase 0 hazard test is therefore described as developed on preliminary data, evaluated on the remaining seeds and replicated on M2 (§5.1), and not as pre-registered.

\textbf{Follow-up experiments (seal nine).} The pre-registration of experiments R and L and the code that reads them were committed before any of their episodes was run, and the reading code before the data were complete. Experiment R finished before experiment L, and R was read early, three times, while L was still being collected: on 27 September 2026 at about 20:30 (Paris time) by the author's coding assistant, at the author's request, when L had 1 of 20 seed pairs; at about 21:30 by the coordinating reviewer, with the same reading function, when L had 5 of 20; and at about 23:00, when the R section of this draft was written, with L at 10 of 20. L was not read before it was complete. Neither the protocol nor the collection of L was changed, and L ran to its pre-registered size.

\textbf{Experiment S (seal ten and its amendment).} The pre-registration, the bench code, the launcher and the reading code were committed before the first episode; an amendment, sealed before launch, replaced the re-run of the bare condition by the August bare episodes (descriptive only) and added power and sleep logging. The launcher waited for another job on the machine to finish before starting. Episodes crossed by a machine sleep were detected (wall clock against a monotonic clock), discarded and re-run from the start (\NpBAinvalidsleep{} episodes). Progress was followed from the launcher's log only; no result was read before the last episode, and the sealed reading was then run once. As a determinism check (D and D1 are identical up to the first counter error): the first counter-error step matched between D and D1 in all \NpBAddBerrmatch{} seeds with an error; neither condition produced a counter error in the remaining \NpBAddBnone{} seeds.

\textbf{The sixteen measurement rules.} Each rule was paid for by a dated error of our own (full text, in French, in \texttt{00\_docs/REGLES\_MESURE.md}); rules 11 and 15 are statements about this protocol, not general laws.

\begin{enumerate}
  \item \textbf{The instrument is younger than the model.} When they disagree, suspect the instrument first.
  \item \textbf{No format that cannot be parsed field by field.} Required prefixes are optional when reading.
  \item \textbf{An ambiguous rule measures which reading the model chose, not drift.}
  \item \textbf{The seal applies word for word, in both directions.} Never relax a criterion after seeing the data --- nor tighten it; reservations are reported next to the verdict.
  \item \textbf{Absence of evidence is not evidence of absence.}
  \item \textbf{The hazard cutoff is pre-registered per substrate}, on a separate exploration set.
  \item \textbf{Show the curve, never the p-value alone.}
  \item \textbf{No contention on the machine during a measurement.}
  \item \textbf{The history keeps the model's real answer}, never the expected one.
  \item \textbf{Injected blocks decorate the current message; the bare message enters the history.}
  \item \textbf{In this protocol, every disruptive block tested degraded the counter, and the block resembling the ledger degraded it most} (reformulated after the irrelevant control contradicted a first, content-based statement). The rule covers disruptive blocks only: the exact state reminder of experiment S is also an injected block, and under it no counter error was observed.
  \item \textbf{Before comparing substrates, check that the prompt scaffold is identical.}
  \item \textbf{A shared inference server is not a substrate.}
  \item \textbf{Measure the noise floor on the statistic you read.}
  \item \textbf{In this protocol, with ollama, unloading the model between episodes made the measurement exactly deterministic.}
  \item \textbf{No sweep of more than 30 GPU-minutes without a passed, dated positive control.}
\end{enumerate}

\textbf{Seals.}

\begin{table}[t]
\centering
\small
\resizebox{\linewidth}{!}{%
\begin{tabular}{lll}
\toprule
document & seal (sha256, first 16) & covers \\
\midrule
\texttt{PREENREGISTREMENT\_phase0.json} & \texttt{f6d99f0dff02280a} & Phase 0 bench, substrates, decoding \\
\texttt{PREENREGISTREMENT\_phase0\_hasard.json} & \texttt{fd6ff9b5983e704d} & hazard falsifier (cutoff 200) \\
\texttt{PREENREGISTREMENT\_etude4b.json} & \texttt{426f203cfc73584d} & 4B study \\
\texttt{PREENREGISTREMENT\_phase1.json} & \texttt{e98eec0efe62dba0} & arms B and D, repair criterion \\
\texttt{PREENREGISTREMENT\_phase1\_brasA.json} & \texttt{2fdc3a4509350a50} & arm A \\
\texttt{PREENREGISTREMENT\_gradation.json} & \texttt{b9caff948036a975} & size map \\
\texttt{PREENREGISTREMENT\_barreau2\_charge.md} & \texttt{11af75cfc235fdb0} & longer-horizon rung (not run) \\
\texttt{AMENDEMENT\_1\_barreau2.md} & \texttt{da14a6720edf9518} & censoring and common cutoff \\
\texttt{PREENREGISTREMENT\_P8.json} & \texttt{6a9f6ef3ff2d5c0b} & follow-up experiments R and L (§5.3--5.4) \\
\texttt{PREENREGISTREMENT\_P10.json} & \texttt{93bcdce57d27f3c0} & follow-up experiment S (§5.5) \\
\texttt{PREENREGISTREMENT\_P10\_amendement1.json} & \texttt{25662909b742e095} & S: bare reference from August, power and sleep logging \\
\bottomrule
\end{tabular}}
\end{table}

\section{Multiple comparisons (Holm, threshold 0.01)}

All \Nholmm{} p-values cited in the paper, excluding the exploratory check of §5.9 and the follow-up experiments of §5.3--5.5, which have their own pre-registered families, in increasing order. \emph{Reference}: one family (all tests). \emph{Per claim}: one family per claim, declared after the reference correction was computed (secondary). \textbf{no} in bold marks a reading significant before correction that does not survive. All tests are one-sided exact Fisher tests except the per-seed Mann-Whitney tests.

\begin{table}[t]
\centering
\small
\resizebox{\linewidth}{!}{%
\begin{tabular}{lllll}
\toprule
test & claim & p & survives reference correction & survives per-claim correction \\
\midrule
step-level 8B > 32B, first fault, common cutoff & C5 & \NffcomIbDCbp{} & \NholmgffcomIbDCbp{} & \NholmfffcomIbDCbp{} \\
per-seed hazard 8B > 32B & C5 & \NseedIbDCbp{} & \NholmgseedIbDCbp{} & \NholmfseedIbDCbp{} \\
per-seed hazard 8B > 14B (common cutoff) & C5 & \NseedIbBEbp{} & \NholmgseedIbBEbp{} & \NholmfseedIbBEbp{} \\
step-level 8B > 14B, first fault, common cutoff & C5 & \NffcomIbBEbp{} & \NholmgffcomIbBEbp{} & \NholmfffcomIbBEbp{} \\
size map, 8B R2 rises (sealed, own cutoff) & C5 & \NsealedIbrCp{} & \NholmgsealedIbrCp{} & \NholmfsealedIbrCp{} \\
R2 rises across step 200, qwen3:8b, M2 & C1 & \NmCrCp{} & \NholmgmCrCp{} & \NholmfmCrCp{} \\
R2 rises across step 200, qwen3:8b, M1 & C1 & \NmBrCp{} & \NholmgmBrCp{} & \NholmfmBrCp{} \\
R2: B below D, M1 & C3 & \NmBbdrCp{} & \NholmgmBbdrCp{} & \NholmfmBbdrCp{} \\
counter: B worse than D, M2 & C2 & \NmCbdctrp{} & \NholmgmCbdctrp{} & \NholmfmCbdctrp{} \\
more 32B seeds at the extremes than 8B & C6 & \NextIbDCbp{} & \NholmgextIbDCbp{} & \NholmfextIbDCbp{} \\
R2 rises within B, M1 & C3 & \NmBarmbrCrisep{} & \NholmgmBarmbrCrisep{} & \NholmfmBarmbrCrisep{} \\
R2 rises within B, M2 & C3 & \NmCarmbrCrisep{} & \NholmgmCarmbrCrisep{} & \NholmfmCarmbrCrisep{} \\
R2: B below D, M2 & C3 & \NmCbdrCp{} & \NholmgmCbdrCp{} & \NholmfmCbdrCp{} \\
R2: B below bare, M1 & C3 & \NmBbArCp{} & \NholmgmBbArCp{} & \NholmfmBbArCp{} \\
counter: B worse than D, M1 & C2 & \NmBbdctrp{} & \NholmgmBbdctrp{} & \NholmfmBbdctrp{} \\
R2: B below bare, M2 & C3 & \NmCbArCp{} & \NholmgmCbArCp{} & \NholmfmCbArCp{} \\
per-seed hazard 14B > 32B & C5 & \NseedBEbDCbp{} & \NholmgseedBEbDCbp{} & \NholmfseedBEbDCbp{} \\
more 14B seeds at the extremes than 8B & C6 & \NextIbBEbp{} & \NholmgextIbBEbp{} & \NholmfextIbBEbp{} \\
R2: D above bare, M2 (claimed null) & C3 & \NmCdArCpworse{} & \NholmgmCdArCpworse{} & \NholmfmCdArCpworse{} \\
R2 rises, 4B, M1 & C1 & \NbErCp{} & \NholmgbErCp{} & \NholmfbErCp{} \\
R1: B below D, M1 (claimed null) & C3 & \NmBbdrBp{} & \NholmgmBbdrBp{} & \NholmfmBbdrBp{} \\
R1: B below D, M2 (claimed null) & C3 & \NmCbdrBp{} & \NholmgmCbdrBp{} & \NholmfmCbdrBp{} \\
R1 rises across step 200, M1 (claimed null) & C1 & \NmBrBp{} & \NholmgmBrBp{} & \NholmfmBrBp{} \\
R1: B below bare, M2 & C3 & \NmCbArBp{} & \NholmgmCbArBp{} & \NholmfmCbArBp{} \\
\bottomrule
\end{tabular}}
\end{table}

\section{Audit: scoring replaced}

The bench's original evaluator gave each line one exclusive label (R3, format, R2, counter, R1, in that order). After an external review showed that this could hide lower-ranked errors, every step was re-scored component by component, and that scoring replaced the original one everywhere in the paper. We document the correction here; the replaced values are not results.

\textbf{Decision rule, committed before the re-scoring was run} (repository commit \texttt{4c1eae5}, after the original data had been seen): the counter result holds if, for the irrelevant block D on both machines, every seed has a counter error against none in the bare arm, and the upper bound of the 95\% seed-bootstrap interval of the R1 error-rate ratio D/bare is below 2; the label "lower bound > 1" if the lower bound exceeds 1; "not conclusive" otherwise. Outcome: \textbf{\NrencCverdict{}}. The body of the paper states the result as a difference of magnitude (§5.2), not as the absence of any effect on local processing.

\textbf{Replaced values.} Under the exclusive labels, the R1 hazard of the injected arms was biased downward: on M2, \NauditmCarmArB{} (bare), \NauditmCarmdrB{} (D) and \NauditmCarmbrB{} (B), against \NmCarmArB{}, \NmCarmdrB{} and \NmCarmbrB{} with the component-wise scoring (hazards censored at the first R1 error; Table \ref{tab:arms} reports the uncensored R1 error rate, \NrenmCarmArB{}, \NrenmCarmdrB{} and \NrenmCarmbrB{}). A test "R1: B below bare" that the exclusive labels had given as significant on M2 (p = \NauditmCbArBpr{}; M1 \NauditmBbArBpr{}) does not hold once re-scored (p = \NmCbArBp{}; Appendix B) and is not a result of this paper. R2 is identical under both scorings; counter hazards differ by at most a few steps at risk.

\begin{table}[t]
\centering
\small
\resizebox{\linewidth}{!}{%
\begin{tabular}{lllllll}
\toprule
machine & arm & counter errors / lines & R1 errors / applicable steps & R1 ratio to bare (95\% interval) & decision rule & R1 errors hidden by the exclusive label \\
\midrule
M1 & 0 & \NrenmBarmActrf{} / \NrenmBarmActrn{} & \NrenmBarmArBf{} / \NrenmBarmArBn{} & --- & --- & \NrenmBarmAhidden{} \\
M1 & D & \NrenmBarmdctrf{} / \NrenmBarmdctrn{} & \NrenmBarmdrBf{} / \NrenmBarmdrBn{} & \NrenmBarmdrBratio{} (\NrenmBarmdrBlo{}--\NrenmBarmdrBhi{}) & \NrenmBarmdrBstate{} & \NrenmBarmdhidden{} \\
M1 & B & \NrenmBarmbctrf{} / \NrenmBarmbctrn{} & \NrenmBarmbrBf{} / \NrenmBarmbrBn{} & \NrenmBarmbrBratio{} (\NrenmBarmbrBlo{}--\NrenmBarmbrBhi{}) & \NrenmBarmbrBstate{} & \NrenmBarmbhidden{} \\
M1 & A & \NrenmBarmactrf{} / \NrenmBarmactrn{} & \NrenmBarmarBf{} / \NrenmBarmarBn{} & \NrenmBarmarBratio{} (\NrenmBarmarBlo{}--\NrenmBarmarBhi{}) & \NrenmBarmarBstate{} & \NrenmBarmahidden{} \\
M2 & 0 & \NrenmCarmActrf{} / \NrenmCarmActrn{} & \NrenmCarmArBf{} / \NrenmCarmArBn{} & --- & --- & \NrenmCarmAhidden{} \\
M2 & D & \NrenmCarmdctrf{} / \NrenmCarmdctrn{} & \NrenmCarmdrBf{} / \NrenmCarmdrBn{} & \NrenmCarmdrBratio{} (\NrenmCarmdrBlo{}--\NrenmCarmdrBhi{}) & \NrenmCarmdrBstate{} & \NrenmCarmdhidden{} \\
M2 & B & \NrenmCarmbctrf{} / \NrenmCarmbctrn{} & \NrenmCarmbrBf{} / \NrenmCarmbrBn{} & \NrenmCarmbrBratio{} (\NrenmCarmbrBlo{}--\NrenmCarmbrBhi{}) & \NrenmCarmbrBstate{} & \NrenmCarmbhidden{} \\
M2 & A & \NrenmCarmactrf{} / \NrenmCarmactrn{} & \NrenmCarmarBf{} / \NrenmCarmarBn{} & \NrenmCarmarBratio{} (\NrenmCarmarBlo{}--\NrenmCarmarBhi{}) & \NrenmCarmarBstate{} & \NrenmCarmahidden{} \\
\bottomrule
\end{tabular}}
\end{table}

The first counter error of a seed moves between the two scorings in \NrenmBarmdfirstctrchanged{} D seed and \NrenmBarmafirstctrchanged{} A seed on M1, \NrenmCarmdfirstctrchanged{} D and \NrenmCarmafirstctrchanged{} A seeds on M2, and in no bare or B seed. The sealed R2 test is unchanged by the re-scoring (p = \NrenmBcBp{} on M1, \NrenmCcBp{} on M2), since only R3 and format outranked R2.

\end{document}